\documentclass{article}

\usepackage[preprint, nonatbib]{neurips_2020}

\usepackage[utf8]{inputenc} % allow utf-8 input
\usepackage[T1]{fontenc}    % use 8-bit T1 fonts
\usepackage{hyperref}       % hyperlinks
\usepackage{url}            % simple URL typesetting
\usepackage{booktabs}       % professional-quality tables
\usepackage{amsfonts}       % blackboard math symbols
\usepackage{nicefrac}       % compact symbols for 1/2, etc.
\usepackage{microtype}      % microtypography
\usepackage{graphics}
\usepackage{graphicx}
\usepackage{eqnarray, amsmath}
\usepackage{algorithm, algorithmicx}
\usepackage{algpseudocode}
\usepackage{xcolor}
\usepackage{caption}
\usepackage{subcaption}
\usepackage{multirow}
\usepackage{tikz}

\title{Clinical trial site matching with improved diversity using fair policy learning }

\author{%
  Rakshith S Srinivasa\thanks{This work was conducted while author was an employee of IQVIA Inc. Correspondence email: rakshith.sharma.s[at]gmail.com}\\IQVIA
   \And
  Cheng Qian  \\ IQVIA
  \AND
  Brandon Theodorou \\UIUC 
  \And
  {Jeffrey Spaeder} \\ IQVIA
  \And
  {Danica Xiao} \\Amplitude
\And {Lucas Glass} \\IQVIA
\And {Jimeng Sun} \\ UIUC 
}
\newcommand{\R}{\mathbb{R}}

\newcommand{\E}{\mathbb{E}}
\newcommand{\vct}[1]{\mathbf{#1}}

\newcommand{\mtx}[1]{\mathbf{#1}}
\newcommand{\vd}{\vct{d}}
\newcommand{\ve}{\vct{e}}

\newcommand{\vh}{\vct{h}}

\newcommand{\vn}{\vct{n}}

\newcommand{\vt}{\vct{t}}

\newcommand{\vv}{\vct{v}}

\newcommand{\mP}{\mtx{P}}

 \def \endprf{\hfill {\vrule height6pt width6pt depth0pt}\medskip}

\begin{document}

\maketitle

\begin{abstract}
 The ongoing pandemic has highlighted the importance of reliable and efficient clinical trials in healthcare. Trial sites, where the trials are conducted, are chosen mainly based on feasibility in terms of medical expertise and access to a large group of patients. More recently, the issue of diversity and inclusion in clinical trials is gaining importance. Different patient groups may experience the effects of a medical drug/ treatment differently and hence need to be included in the clinical trials. These groups could be based on ethnicity, co-morbidities, age, or economic factors. Thus, designing a method for trial site selection that accounts for both feasibility and diversity is a crucial and urgent goal. In this paper, we formulate this problem as a ranking problem with fairness constraints. Using principles of fairness in machine learning, we learn a model that maps a clinical trial description to a ranked list of potential trial sites. Unlike existing fairness frameworks, the group membership of each trial site is non-binary: each trial site may have access to patients from multiple groups. We propose fairness criteria based on demographic parity to address such a multi-group membership scenario. We test our method on 480 real-world clinical trials and show that our model results in a list of potential trial sites that provides access to a diverse set of patients while also ensuing a high number of enrolled patients.
\end{abstract}

\section{Introduction}

The ongoing pandemic has highlighted the importance of reliable clinical trials in healthcare. For successful clinical trials, it is crucial to ensure two outcomes: i) timely and ethical enrollment of subjects who conform to the inclusion and exclusion criteria \footnote{All clinical trials specify certain \textit{inclusion criteria} which describe a set of conditions to be met by patients to be enrolled in the trial and \textit{exclusion criteria}, which describe conditions that determine patients who cannot be enrolled in the study.}, and increasingly ii) enrollment of a cohort of subjects (study population) that enables a characterization of the safety and efficacy of the intervention in a population that corresponds to the characteristics of the overall population (treatment population).  Such alignment between study population and treatment population is essential for the concept of fairness in clinical research.  Due to such considerations, selecting sites (a site is a medical facility where a clinical trial is conducted) to conduct a given clinical trial is a complex task. 

Achieving an alignment between the study population and the overall population requires improving the diversity of the study population.  Improving diversity in clinical trials is an important issue that is now gaining attention from the clinical trials community.  The U.S Food and Drug administration (FDA) recently issued guidelines on addressing diversity in clinical trials \cite{fda_div}. As stated in the document, the FDA ``recommends approaches that sponsors of clinical trials intended to support a new drug application or a biologics license application can take to increase enrollment of underrepresented populations in their clinical trials". In this paper, we propose a method to directly address this problem.

Our goal is to use rich historical hospital data, investigator performance data from past clinical trials, and patient demographics at sites along with machine learning to address the complex problem of trial site matching in way that address both enrollment and diversity targets. For a new clinical trial, contract research organizations (CROs) select investigators from a large existing pool using a set of pre-defined criteria. In this paper, we seek to improve this selection process by learning the complex relationships between investigators and their performance in clinical research using multi-modal data. 

To formalize the problem, we pose the task of trial site matching as a fair ranking problem, where we rank the list of potential trial sites in order to maximize expected patient enrollment and patient diversity. In the subsequent text, we use the terms `investigator' and `site' interchangeably for convenience. Towards this end, we propose a machine learning algorithm that learns to generate a Top-$K$, trial-specific ranking of a given list of investigators. Given a new clinical trial and a list of $M$ investigators ($M > K$), the algorithm selects the \textit{top K} investigators from this list. The set of Top-$K$ investigators are chosen by simultaneously optimizing for patient enrollment and patient diversity.  Both $K$ and $M$ are user-specified parameters and do not influence the algorithm design. To account for the two different goals of effective trials described above, we use a \textbf{learning-to-rank} framework with \textbf{fairness constraints}. Given a new trial/ study, we use a learned model to rank a set of potential sites/ hospitals such that the Top-$K$ sites together provide access to a large and diverse patient cohort.

For the scope of this paper, we define diversity based on the race and ethnicity of the patients. In particular, we assume the following groups: White, Hispanic, African-American, Asian, Mixed, and Others. Note that our method works for any definition of groups/ ethnicity. We chose these groups based on the population data that we have access to.  We define a diverse patient cohort as one that reflects the representation of these groups in the underlying population. Our goal is to develop an algorithm that chooses such a patient cohort across a set of $K$ investigators. Note that an individual investigator selected by our algorithm to be in the Top-$K$ may not meet this criterion, but together, the Top-$K$ set of investigators will. Further, we note that the definitions of groups and the definition of diversity can be changed to reflect other considerations such as age, co-morbidity, and nationality-based groups (among others).

In our setup, we associate each investigator with two sets of features: i) their access to patients relevant to a given clinical trial ii) the demographics of the patient population the investigator has access to. Hence, unlike standard settings in the fairness literature where each data point is associated with a \textit{single group}, each investigator has a \textbf{mixed} representation comprising of \textbf{all the groups}. We refer to this as \textbf{multi-group membership}. In the sequel, we demonstrate how the multi-group membership demands new considerations in algorithm design and then present a method to address this challenge.

Another technical challenge in developing a trial site matching algorithm is the complex nature of the associated data. We use multi-modal data sources to build rich representations of clinical trials and investigators that include unstructured text data, trial phase, trial type, International Classification of Diseases (ICD)-10 codes of the conditions targeted by the trial and a list of ICD-10 codes that investigators have treated in the past. We use a deep learning based ranking method to capture the information from these different data sources. Using a deep learning based framework allows us to generate useful embeddings to represent both clinical trials and investigators. 

For our experiments, we use real-world clinical trial data from IQVIA's clinical trial enrollment database to evaluate our algorithm. We show that without any diversity based constraints, our proposed method  learns to identify investigators who are well suited for the trial in terms of patient access. Once diversity constraints are added, we show that our model achieves a trade-off to account for diversity. We show that our model achieves a better demographic composition by choosing a different set of investigators without a large decline in expected patient enrollment. 

\section{Related work}

Clinical trials form the backbone of medical practice. Very often, they take many years to be completed and require huge sums of monetary investment. For any clinical trial, choosing the right set of investigators for the trial often becomes crucial in reaching the patient enrollment goals. Hence, trial site matching is inherently a ranking problem. However, it is still common that this problem is often solved manually or using very basic matching between a trial's target condition and investigators' treatment history statistics \cite{site_sel_chong}. Further, increasing diversity in the recruited patient cohort is now a priority for many CROs. This paper seeks to utilize a rich set of multi-modal datasets and provide a machine learning-based solution to these problems. 

\textbf{Learning to rank}

Our work is naturally related to the ranking literature from information retrieval. Literature on learning to rank consists of two types of problem setups: first, where the relevance or ``score" of items to be ranked is available \cite{fa*ir, asudeh19} and second, where the scores need to be learnt. Our work is directly related to the second scenario. Some approaches to that second problem include \cite{softrank, Listnet, adarank, rankNet}. Authors in \cite{rankNet} consider a pair-wise ranking problem where the training data consists of pair-wise comparisons. We consider a list-wise approach, where a full list of items need to be ranked at once. While \cite{softrank, Listnet, adarank} consider a similar listwise approach, their frameworks cannot incorporate general ranking based cost functions due to the non-differentiability of such functions. In our approach, we use a policy learning based framework similar to \cite{singh2019ranking}, which can accommodate any ranking based loss functions. 

\textbf{Fairness in machine learning and ranking}

Fairness in machine learning is becoming an increasingly important research topic \cite{barocas-hardt-narayanan}. As machine learning algorithms become a part of many real-life systems such as candidate selection, clinical trial site selection, banking and financial applications, it is crucial to ensure that these algorithms do not have a disparate impact on certain groups or communities.

Fairness being an important topic, there are many proposed solutions to ensure that algorithms are fair. Most of these solutions are in the context of classification and regression, as these are very common goals of most machine learning methods. For a full survey of the literature, see \cite{caton2020fairness, beutel2019fairness}. Most fairness literature follows the following model: each data point belongs to a group or has some sensitive attributes; a fair algorithm is designed such that the group identity of the sensitive attributes does not contribute to the outcome. We consider the learning-to-rank problem and aim to develop a fair ranking approach for trial site matching.

Our method for trial site matching is mainly derived from the learning to rank framework proposed in \cite{singh2019ranking}. The authors in \cite{singh2019ranking} first proposed a policy learning based framework to learn ranking of items. Their method allows for the usage of any ranking-based loss function due to the usage of a probabilistic ranking model and policy learning.  We use their policy learning framework as a building block for our Top-$K$ ranking framework. However, a key difference in our work is our approach to incorporate fairness into ranking. In particular, the authors in \cite{singh2019ranking} address the problem of disproportionate exposure of items with a high relevance, where exposure is defined using the classical definition from ranking literature. This leads them to impose a one-sided fairness constraint that ensures that items with lower relevance or score are not excessively penalized. Such a one-sided constraint, while being applicable to the singular group membership scenario, breaks down when we consider multi-group membership. We provide more details in Section \ref{subsec:fair_setup}.

\textbf{Clinical trial recruitment, machine learning and diversity}

Clinical trials  recruit investigators to participate in clinical trials and studies. This is part of the development cycle of drugs and medical interventions and is usually outsourced by pharmaceutical companies to CROs.

More recently, machine learning is being used in various capacities by the drug development and clinical trials industry \cite{MLCT_covid}. In \cite{vanderbor17}, the authors use responses to a questionnaire to predict investigator enrollment performance in Phase 3 clinical trials. In \cite{Gligorijevic}, the authors propose to first generate embeddings of multi-modal data associated with clinical trials and then use an inner product based aggregation to recommend investigators for trials. We use an end-to-end framework that is trained specifically for ranking. Authors in \cite{doctor2vec} consider predicting the enrollment rates of investigators for a given trial. In this paper, we go beyond regressing enrollment rates and provide a ranking of the investigators.

In addition to methodological differences from the above works, this paper also addresses another important concern: diversity in clinical trial recruitment.  Clinical interventions may have varying outcomes across different patient subgroups based on race, ethnicity and socio-economic factors, as noted in \cite{Clark2019increasing, knepper2018will, burchard2014medical}. Further, clinical interventions are often based on quantitative characterizations of disease severity, that may not correspond to actual disease severity \cite{eberly2018psychosocial, Pierson2020pain}. In such cases, it is important for clinical trials to evaluate interventions across various patient subgroups. Our method directly addresses these concerns by developing a machine learning framework to ensure diversity in clinical trial recruitment.

\section{Fairness in ranking for trial site matching}

\subsection{Problem statement:}

Assume a list of $M$ potential trial sites or investigators $\{ \vd_1, \cdots, \vd_M \}$. Each site/ investigator  $\vd_i$ is characterized by their medical expertise and specialty areas. Further, we assume that each site provides access to a set of patients, each of whom belongs to one of $L$ groups. For example, these could be ethnicity-based sub-populations. Hence, each site $\vd_i$ is associated with a \textit{distribution} of patients over the $L$ groups. We represent these distributions for all the sites together using matrix $\mP$ of size $M \times L$. Here, $\mP[i,l]$ represents the percentage of the patients at site $i$ that belong to group $l$. Given a new trial $\vt$ and list of $M$ sites, our goal is to provide a ``Top-$K$" ranking, a list of $K$ ($K < M$) sites that together maximize the suitability of the sites for the trial and ensure demographic parity between the $L$ groups. Note that our method can generalize to previously unseen sites.

In order to do so, we leverage patient enrollment data for past clinical trials conducted at these sites. Specifically, we use past trials ${\vt_1, \vt_2, \cdots \vt_N}$, where for each trial $\vt_i$, we have an associated vector $\ve_i \in \R^M$ that provides patient enrollment rates at the $M$ sites.  The enrollment rate captures the ability of a site to enroll patients for a given trial and is defined as follows:
\begin{equation}
    \text{enrollment rate} = \frac{\text{Number of patients completing the trial}}{\text{enrollment window in months}} .
\end{equation} This ability itself is characterized by latent factors such as medical expertise area, location and access to patients. Using the enrollment rates as measures of suitability between a trial $\vt$ and a site $\vd_i$, we train a model to learn a ranking policy $\pi: \vt \rightarrow r$, where $r$ is a ranking (permutation) on the $M$ sites. During inference, given a trial $\vt$ and a list of $M$ sites, our trained model can be used to obtain the Top-$K$ sites. We explain the details of our model below.

\begin{figure}
    \centering
    \hspace*{-0.5in}\includegraphics[scale=0.5]{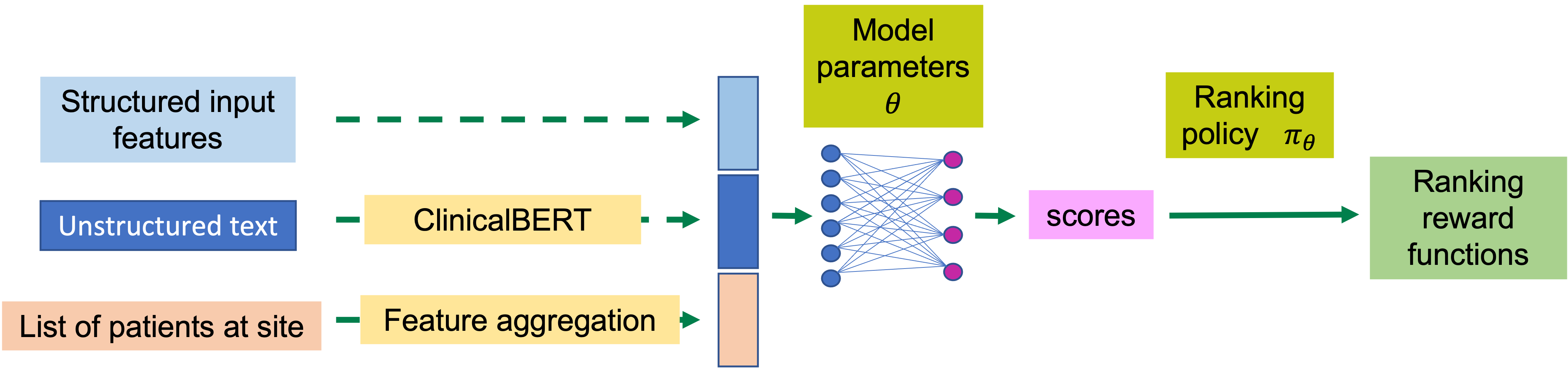}
    \caption{Schematic representation of our model. Given a trial description $\vt$ and a collection of potential sites $\{ d_1, \ \cdots, \ d_M \}$, we predict the K most suitable sites for the trial by in terms of suitability and patient diversity. }
    \label{fig:model}
\end{figure}
\subsection{Fair Top-$K$ ranking using policy learning}

We use a policy learning based framework to learn the Top-$K$ selection from a given list. In particular, our model learns to assign a score to each site in the input list and chooses the Top-$K$ sites based on the scores. A policy then refers to the set of rules used to select the Top-$K$ sites. Note that a simple policy is to just sort the scores and choose K sites with the highest scores. However, using such a simple deterministic policy reduces the model's ability to learn during training, as shown by our experiments later in the paper. A randomized policy learning framework for learning to rank was proposed in \cite{singh2019ranking}. Their framework provides an elegant way to incorporate ranking based loss functions directly into model training. In our work, we develop on this framework to account for multi-group membership. In particular, while the framework proposed in \cite{singh2019ranking} is useful for ranking in general and for fair ranking under the singular-group membership constraint, we show in Section \ref{subsec:fair_setup} that it cannot be applied to the multi-group scenario.  We then address this gap by developing a fairness metric that is applicable to the multi-group membership. We outline the various modules of our proposed model below. 

\subsubsection{Embedding learning}

The first task in our ranking model is to learn how to score the candidate sites for a given trial. We learn a model to map a trial $\vt$ to a set of relevance scores $\vh \in \R^M$ that captures the suitability of the sites to a trial.
This can be represented as
\begin{equation}
\label{eq:score_model}
    \vh(\vt) \in \R^M = f_\theta (\vt, \{ \vd_1, \ \cdots, \vd_M \}).
\end{equation}
where $f_\theta$ takes the form of a neural network.

\subsubsection{Probabilistic ranking policy}

As described earlier, the score vector $\vh(\vt)$ will be used by the policy to generate a ranking of the sites. Similar to \cite{singh2019ranking}, 
given the score vector $\vh(t)$, we use the Plackett-Luce model \cite{plackett75permutations} to define a probabilistic policy in the space of rankings (permutations) of the sites. Let $r$ be a permutation of the sites $1$ to $M$ and let $r[i]$ denote the index of the site at the $i$th rank. Then, the probability of $r$ (w.r.t the set of all possible rankings) is given by
\begin{equation}
    \pi_\theta(r | \vt) = \prod_{i = 1}^M \frac{\exp(\vh(\vt)[r(i)])}{\sum_{j = i}^M \exp(\vh(\vt)[r(j)])}.
    \label{eq:rank_policy}
\end{equation} As stated in \cite{singh2019ranking}, we can sample a ranking in the following way: starting from the top, documents
are drawn recursively from the probability distribution resulting from Softmax over the scores of the
remaining documents in the candidate set, until the set is empty. 

Given a ranking sample, our algorithm requires computing the probability of the sample according to Equation \eqref{eq:rank_policy}.
Sampling from the above distribution can be performed efficiently for small values of $M$. However, for larger values of $M$, we find empirically that sampling from \eqref{eq:rank_policy} and computing the probability for each sample is expensive (especially while drawing many Monte-Carlo samples, as explained in the sequel) and also unstable. Hence, we use a simple proxy defined as
\begin{equation}
\label{eq:proxy_policy}
    \widehat{\pi}_\theta (r|\vt) = \sum_{k=1}^K \frac{\exp(\vh(\vt)[r(i)])}{\sum_{j = 1}^M \exp(\vh(\vt)[r(j)])}
\end{equation}
to compute the ``Top-$K$ probability" of a given ranking sample. Since this is just given by a simple sum, computing the probability of Monte-Carlo ranking samples is efficient.

% \subsection{{\color{red}Probability of Top-$K$ ranking}}

\subsubsection{Utility of a ranking}

In the training phase of our method, the model needs to learn to generate good rankings. Sites with high enrollment in the past should be assigned higher scores, resulting in them being ranked higher. To encourage such a learning process, policy learning assigns a reward or a utility function for each ranking. The model parameters are then optimized such that rankings with higher rewards are preferred over those with lower rewards. 

One of the main advantages of using a policy learning based framework for ranking is that one can directly optimize over even non-smooth utility functions. This is highly useful for learning to rank due to the ordinal nature of rankings. This was first proposed in \cite{singh2019ranking}. However, unlike in \cite{singh2019ranking}, we are interested in the Top-$K$ items and not in the entire ranking space. We define the utility of a ranking policy $\pi$ for trial t as 
\begin{equation}
\label{eq:utility}
   U(\pi|\vt) =  \mathbb{E}_{r \sim \pi } [\ell(r, \ve_t)]
\end{equation} where $\ell(\cdot)$ is a suitable reward function. To focus on the Top-$K$ ranked items, we define $\ell (\cdot )$ as below:
\begin{equation}
    \label{eq:util_loss}
    \ell(r, \ve_t) = \sum_{k = 1}^{K} \ve_t[k] - \sum_{k = K+1}^{M}\ve_t[k].
\end{equation}
Maximizing the above utility function directly leads to maximizing the expected number of enrolled patients in the trial. The second term in Equation \eqref{eq:util_loss} discourages having sites with high enrollment rates outside of the Top-$K$ ranking.

\begin{figure}
    \centering
    \includegraphics[scale=0.5]{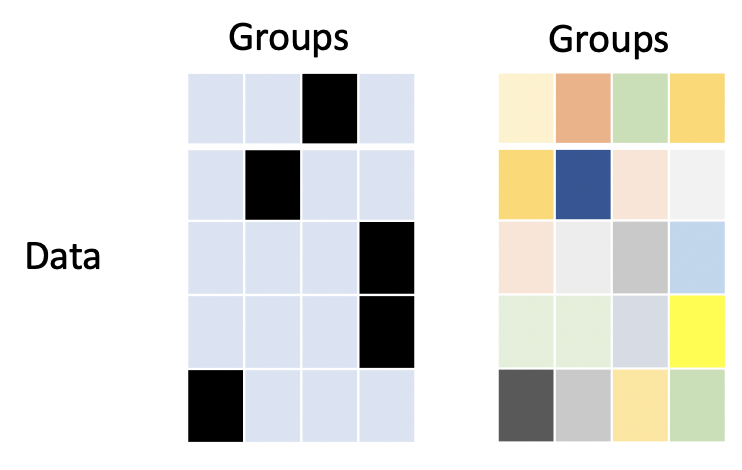}
    \caption{Each site/ hospital is associated with multiple groups, unlike in traditional fairness considerations where each data point is assumed to belong to a single group. Further, knowledge of the multi-group membership matrix may be noisy.}
    \label{fig:multi-group}
\end{figure}

\subsection{Fairness constraints for multi-group items}

Fairness is becoming an increasingly important concern in machine learning. It is of importance in the healthcare industry as well. Any biases in models can have a significant impact on the nature of healthcare available to different groups and communities. For the purposes of this paper, we define fairness as equivalent to diversity in the patient cohort selected for a clinical trial. Accordingly, we highlight an important difference between general-purpose fairness considerations and as it applies to our problem of ranking trial sites. We then propose an entropy-based mechanism to encourage site ranking models to be fair.

In general, fairness in machine learning assumes that each data point belongs to one of many groups. For example, the groups could be based on gender, race, income category, etc. The general framework to ensure fairness then requires that the risk of a model with parameters $\theta$,  $\mathbb{R}(\theta, X, Y)$ is similar (in a pre-defined sense) to the risk of the model on each group $\mathbb{R}(\theta, X, Y : X \in G_j)$ for all $j$, where $G_j$ is the $j$th group. In other words, the group membership of each data point is assumed to be ``singular": if a group $j$ is associated with a data point $X$, no other group is associated with the same data point. 

In this paper, the items (trial sites) are each associated with \textit{distribution} across \textbf{all} the groups. For instance, each site/ hospital has access to a diverse set of patients. Therefore, each site itself does not belong to any single group. We illustrate this difference between the standard paradigm of fairness and the problem under consideration in Figure \ref{fig:multi-group}. Existing methods to generate fair rankings (including \cite{singh2019ranking}) aim to ensure that items with lower scores are not disproportionately affected. This is achieved by imposing ``exposure" based fairness constraints. In the following section, we describe such constraints and illustrate why they are not directly applicable in the multi-group membership scenario. We then propose a framework that addresses this gap.

\subsubsection{Exposure based loss functions for fairness in ranking}
\label{subsec:fair_setup}
There are only a few studies in the ranking literature that study fairness. Further, only a subset of the methods considers the problem of learning to rank, whereas most others assume that the score of each item to be ranked is available. One such work which also considers learning to rank is \cite{singh2019ranking}. Here, the authors use an exposure regulating fairness framework, where exposure is defined as the expected attention an item receives, given a ranking policy. We now elaborate on the exposure-based loss below and show why it is not applicable in our case. 

Let $\vd_i$, $i=1, \cdots, M$ denote the $M$ sites and let $r$ denote a ranking of these sites. Then, the position bias of item $i$, denoted by $\vv_i$ is defined as the expected attention item $i$ receives on average. In our case, we can define this as
\begin{equation}
\vv_r[i] = \begin{cases}
1 & i \in \{r(1), \cdots, r(K) \} \\
0 & \text{otherwise}.
\end{cases}
\end{equation} Then, for a given ranking policy $\pi$, the exposure of item $i$ is defined as 
\begin{equation}
    v_\pi(i) = \E_{r \sim \pi} \vv_r[i].
\end{equation} When each item $i$ belongs to a single group, the exposure of a group is defined as 
\begin{align}
    v^g_\pi(l) = \sum_{i \in \mathcal{G}_l} v_\pi(i).
\end{align} On the contrary, when each item $i$ is associated with a distribution over all the groups, exposure of group $l$ can be defined as 
\begin{align}
    v^g_\pi(l) =  \sum_{i} \mP[i,l]\cdot v_\pi[i] \nonumber \\
\end{align} where $\mP$ is the group-membership matrix described earlier.

To define the exposure-based loss, one also needs to define the merit associated with each group. In case of single-group membership, the merit for each item is just the ground truth score of that item. In our case, the merit of each site is the past patient enrollment $\ve_t[i]$. In the multi-group membership case, the merit of a group may be defined as the average population percentage of that group:
\begin{align*}
    M(l) & = \frac{1}{\sum_i \ve_\vt[i]} \sum_i \mP[i,l] \ve_\vt[i] \\
    & = \sum_i \mP[i,l] \gamma_{\vt, \ i}
\end{align*} where $\gamma_{\vt, \ i} = \frac{\ve_\vt[i]}{\sum_i \ve_\vt[i]}$ is the relative enrollment percentage of site $i$. Note that if $\gamma_i$ has a relatively flat distribution over $i$'s, $M(l)$ is roughly equivalent to the average population percentage of group $l$. 

Exposure based ranking loss function (See \cite{singh2019ranking}) is then defined as
\begin{equation}
\label{eq:os_loss}
    \max \{0, \frac{v^g_\pi (l)}{M(l)} - \frac{v^g_\pi (l')}{M(l')} \}
\end{equation} $\forall \ l, l'$ such that $M(l) > M(l')$ (See for example \cite{singh2019ranking}). Intuitively, this imposes that the exposure received by under-represented groups is not disproportionately low. While the opposite direction also needs to be imposed, explicitly imposing it makes the problem infeasible \cite{singh2019ranking}. In the case of single-group membership, the opposite direction is automatically taken care of by maximizing utility. But in case of multi-group membership, it is not. To see why, consider a case where there are $K$ sites with a large representation of minority groups and have high historical patient enrollment rates. Choosing these $K$ sites maximizes both utility and satisfies the exposure based constraint. However, the majority group is not represented at all and hence the site selection is not diverse. We illustrate the above drawback of exposure based constraints for multi-group membership further, using a hypothetical example with $M=5$, $K=1$ and $l=4$, with details as in Table. \ref{tab:hyp_trial_feat} Note that in this case, choosing site 1 is optimal if the loss in Equation \eqref{eq:os_loss} is used, since it also maximizes the utility defined in Equation \eqref{eq:util_loss}. However, the site that is most representative of the underlying population is site 4. Therefore, while exposure based fairness solutions work well for single-group membership, they are not suitable for multi-group membership.

\begin{table*}[t]
\begin{center}
\caption{An illustration of why exposure based loss functions are not suitable for multi-group membership.}
\label{tab:hyp_trial_feat}
    % \resizebox{\textwidth}{!}{
    \begin{tabular}{c|c|c|c|c|c}
        \toprule
        Site & group 1 & group 2& group 3& group 4 & Enrollment \\
        \midrule
        Site 1 & 0.2 & 0.35 & 0.25 & 0.2 & 100\\
        Site 2 & 0.6 & 0.15 & 0.15 & 0.1 & 100\\
        Site 3 & 0.65 & 0.15 & 0.1 & 0.1 & 100\\
        Site 4 & 0.55 & 0.18 & 0.15 & 0.12 & 100\\
        Site 5 & 0.7 & 0.2 & 0.08 & 0.02 & 100\\
         \midrule
        Avg. percentage & 0.54 & 0.21 & 0.15 & 0.1 
    \end{tabular}
\end{center}
\end{table*}

% \textbf{Fairness based on pre-specified representation criteria}\\
% In this section, we formulate a fairness metric that accounts for uncertain and multi-group membership of the sites. We first formulate the fairness criteria using the noisy group membership matrix $\mP$ and then modify it to include the uncertainty in our knowledge of $\mP$. Let each site $d_i$ be associated with a discrete distribution over the $L$ groups, $\vp_i$ = $\mP(i,:)$. Then, $\vp_i \in \R^{L}$ and $\sum_l \vp_{i}(l) = 1$. For a given ranking $r$, the (normalized) expected number of patients in group $l$ across the Top-$K$ sites, $n_l$ is given by
% \begin{equation}
%     \mathbb{E} \  n_l = \frac{1}{K}\sum_{i=1}^K \vp_{r(i)}(l).
% \end{equation} Taking expectation over all possible rankings, we have 
% \begin{align}
%     \mathbb{E}_{r \sim \pi} n_l & = \mathbb{E}_{r \sim \pi} \frac{1}{K}\sum_{i=1}^K \vp_{r(i)}(l) \\
%     & = \frac{1}{K} \sum_{i = 1}^M \mathrm{Prob}(r(i) \leq K) \vp_i(l)
% \end{align}

% In order to ensure demographic parity, we enforce a lower bound on the expected number of patients from each ethnic group by a pre-defined group-specific constant. Therefore, our fairness criteria is the following 
% \begin{align}
%     \E_{r \sim \pi} n_l & \geq \gamma_l \\
%     \frac{1}{K} \sum_{i = 1}^M \mathrm{Prob}_{r \sim \pi}(r(i) \leq K) p_i(l) & \geq \gamma_l \ \forall \ l = 1, \cdots, \ L
% \end{align} where the probability is over the distribution over rankings  due to the policy $\pi$.

\textbf{Entropy based diversity in clinical trials}\\
We design a framework based on entropy to encourage our model to choose a set of $K$ sites that collectively maximize the diversity of the underlying patient cohort. Let $\vn$ represent any feasible group membership vector ($\vn[i] \geq 0$, $\sum_i \vn[i] = 1$). Then, the entropy of $\vn$ is defined as 
\begin{equation}
    \label{eq:entropy}
    H(\vn) = \sum_i \vn[i] \log \vn[i].
\end{equation} This is the standard definition of entropy from information theory. Note that the more uniform the group representation vector is, the higher the entropy is. The highest entropy is attained by the uniform vector $\vn[i] = 1/L$ for all $i = 1, \ \cdots, \ L$. 
% Further, any particular target distribution can be encouraged by using a KL divergence loss between $\vn$ and the target distribution.

The entropy-based metric can then be directly incorporated into the policy training framework described above. Intuitively, selecting a set of $K$ sites that together have a high entropy results in a higher reward and vice-versa. We can then define a fairness based reward function for a policy as 
\begin{equation}
\label{eq:fairness_reward}
  F(\pi|\vt) =  \mathbb{E}_{r \sim \pi } [\ell_f(r, \ve_t)]
\end{equation}
where
\begin{equation}
    \ell_f(r,\mP) = H\left( \frac{1}{K} \sum_{k=1}^K \mP[r[k],:] \right).
\end{equation}
$F(\pi|t)$ then represents the average entropy generated by sampling rankings from the policy $\pi$. The policy learning framework will then encourage the model to rank sites such that the resulting entropy is high, resulting in a diverse patient cohort.

\subsubsection{Optimization framework}

With the utility and fairness reward functions defined in Equations \eqref{eq:utility} and \eqref{eq:fairness_reward}, we can use the standard policy learning paradigm to set up the following optimization problem:

\begin{eqnarray}
    & \pi^*  = \underset{\pi}{\mathrm{argmax}}\ \E_{t} \ \  U(r|\vt)  + \lambda F(r|\vt)
\end{eqnarray}
A gradient based algorithm can then be used to carry out the above optimization, similar to \cite{singh2019ranking}. We elaborate on the details of this procedure in the next section.

\section{Algorithm implementation and experimental setup}
\label{sec:implementation}
In this section, we provide details regarding the implementation of our proposed framework. To begin with, we restate some of the notations used in the paper and also define some new notations. Our framework takes as input features of a clinical trial, $\vt \in \R^{p}$ and a set of $M$ potential investigators, denoted by $\{ \vd_1, \cdots, \vd_M \}$, where each investigator is represented in $\R^{q}$ as a feature vector. Our goal is to choose the Top-$K$ investigators from this list of length $M$. 

\subsection{Policy network}

For a given trial $\vt$, our model first predicts a score for each of the $M$ investigators. The scores are generated using a deep neural network, denoted as $f_\theta$, where $\theta$ is the set of parameters of the deep neural network. The input to the neural network is the trial features (of dimension $p$), along with the set of investigator features. We denote the set of $M$ outputs of the deep neural network containing the scores as $\vh(\vt) \in \R^{M}$. Hence, our model can be denoted as 
\begin{equation}
    f_\theta: \R^{p + q} \rightarrow \R.
\end{equation} 

\subsection{Deterministic Vs Randomized ranking}

Given the score vector $\vh(t)$, our next step is to compute a ranking of the investigators. This can be done in either a deterministic way or in a probabilistic way. Under deterministic ranking, the scores can be directly converted into an ordering of the investigators by sorting them. However, this could lead to a lack of exploration: even slight differences in score due to noise can cause the algorithm to always prefer a particular ranking over others. For this reason, we use randomized sampling of rankings, as defined in Equation \eqref{eq:rank_policy}. We compare our randomized policy with the deterministic policy as a baseline.
\subsection{Policy optimization using Monte-Carlo sampling}
In order to optimize the neural model $f_\theta$ to learn a mapping from trial-investigator pairs to the score vector, we use gradient-based training of the model. Each ranking of the investigators results in certain utility and diversity measures, as described before. In particular, we carry-out the following optimization: 
\begin{align}
    &\underset{\pi_\theta}{\mathrm{arg max}} \  \sum_{t} \E_{r \sim \pi_\theta}  U(r|t) + \lambda F(r|t) \\
    &\underset{\pi_\theta}{\mathrm{arg max}} \ \sum_{t} \E_{r \sim \pi_\theta}  \left [ \ell (r, \ve_t) \right ] + \lambda \E_{r \sim \pi_\theta}l_f(r, \mP) \\
        &\underset{\pi_\theta}{\mathrm{arg max}} \ \sum_{t} \E_{r \sim \pi_\theta}  \left [ \ell (r, \ve_t) \right ] + \lambda \E_{r \sim \pi_\theta}H(r, \mP) \\
   & \underset{\pi_\theta}{\mathrm{arg max}} \ 
    \sum_{t} \ \  \E_{r \sim \pi_\theta} \left [ \sum_{k=1}^K \ve[r[k]] - \sum_{k' = K+1}^M \ve[r[k']] \right] + \lambda  \E_{r \sim \pi} H(\frac{1}{K} \sum_{k=1}^K \mP[r[k],:] ).
\end{align}Note the more uniform a distribution is, the higher the entropy. Hence, we maximize the entropy of the average distribution over groups associated with the Top-$K$ investigators chosen by the model.

In order to carry-out the above optimization, we use a gradient-based algorithm. This requires us to compute the gradient of the two terms in the above equation:
\begin{align}
  \nabla_\theta U(r|t) & =  \nabla_\theta  \left ( \E_{r \sim \pi_\theta} \left [ \ell (r, \ve_t) \right ]  \right ) \\
   \nabla_\theta F(r|t) & = \nabla_\theta \left ( \E_{r \sim \pi_\theta} H(r, \mP) \right ). 
\end{align} However, computing this expectation exactly is an intractable strategy, as the space of all rankings is very high dimensional. Similar to \cite{singh2019ranking}, we use the REINFORCE trick to switch the order of computing expectation and gradients:
\begin{align}
   \nabla_\theta U(r|t)  =  \nabla_\theta  \left ( \E_{r \sim \pi_\theta} \left [ \ell (r, \ve_t) \right ]  \right ) & = \E_{r \sim \pi_\theta}  \left ( \ell (r, \ve_t)  \cdot  \nabla_\theta \log \pi_\theta (r) \right ) \\
   \nabla_\theta F(r|t) =   \nabla_\theta \left ( \E_{r \sim \pi_\theta} H(r, \mP) \right ) & = \E_{r \sim \pi_\theta}  \left ( H(r, \mP) \cdot  \nabla_\theta \log \pi_\theta (r) \right )
\end{align} where $\pi_\theta (r)$ is the probability of the ranking $r$. Using these gradients, we can update our model as 
\begin{equation}
    \theta \leftarrow \theta + \eta \nabla_\theta \left ( U(r|t) + \lambda  F(r|t) \right)
\end{equation}

\subsection{Data preparation}

In this section, we  provide details about the dataset we use and describe the features we use to train our model.  For data regarding clinical trials, we use IQVIA's internal dataset that contains historical patient enrollment data for 4,264 past clinical trials conducted by a cumulative of 31,836 investigators. For each trial, this dataset provides the total number of patients enrolled by an investigator after randomized assignment to the control groups. This data is available for each investigator who participated in the trial.

\textbf{Clinical trial features}

Each clinical trial in our database is associated with a unique identification number called the NCT ID. The NCT ID can be used to query the details of each clinical trial from \url{clinicaltrials.gov}, which is a government database of private and publicly funded clinical trials. Using this website, we collect the following features for each trial: trial phase, trial type, trial, ICD conditions, eligible ages of patients, and the inclusion/exclusion criteria. These features are formatted as below.
\begin{itemize}
    \item Trial phase: Integer 
    \item Trial type: There are three different types of trials in our dataset: Observational, Interventional, Expanded Access. We use a 4-length one-hot vector to represent this (with one extra for when the trial type is absent from the data).
    \item Inpatient flag: Binary 
    \item ICD-10 code: For each trial, we obtain the condition the trial is seeking to address. We then use the first letter of the ICD-10 code. We represent it as a one-hot vector of length 26, with the entry for the first letter. This is motivated by the fact that the first letter of the ICD-10 code represents the category of the condition.
    \item Minimum and maximum age of eligible patients are in the form of integers. 
    \item Inclusion and Exclusion criteria: Both inclusion and exclusion criteria are text-based features. For each of these, we use the pre-trained clinicalBERT model \cite{alsentzer2019} to extract embeddings.
\end{itemize}
Along with these features, we added two other features that are available in the IQVIA database: primary therapeutic area and whether the trial requires inpatient admission. The format of these features and their sizes are outlined in Table. \ref{tab:trial_feat}.

\begin{table*}[t]
\begin{center}
\caption{Features used to represent clinical trials}
\label{tab:trial_feat}
    % \resizebox{\textwidth}{!}{
    \begin{tabular}{c|c}
        \toprule
        Feature & Format \\
        \midrule
        Therapeutic area & 23-length multi-hot \\
        Inpatient flag & binary\\
        Trial phase & Integer $\in \{1,2,3,4\}$ \\
        Trial Type & 4-length one-hot\\
        Trial ICD-10 code & 26-length one-hot \\
        Min. age & integer \\
        Max. age & Integer \\
        Inclusion critera & 768-length real vector\\
        Exclusion criteria & 768-length real vector\\
         \bottomrule
    \end{tabular}
\end{center}
\end{table*}

\textbf{Investigator features}

For each clinical trial in our database, we have access to a list of investigators who participated in the trials. For each investigator, we have the location information (zip code). Further, we have access to the list of patients who visited the investigator. This gives access to a rich set of features that represent the medical expertise of an investigator along with the details of the patient visit in the form of insurance claims. This in turn dictates the ability of the investigator to reach patients who might be eligible for the trial. For each investigator, we collect all the patient visits during the years 2016 through 2020 years using IQVIA's insurance claims data.  This dataset records each insurance claim made across the entire country of USA. Each claim is further associated with a set of ICD-10 codes that captures the particular medical diagnosis. 

Each ICD-10 code is a combination of letters and digits. For example, the set of conditions related to Tuberculosis are represented by ICD-10 codes starting in A15 followed by further digits/ letters representing particular forms of the condition. Further, ICD-10 codes are ordered such that using the first letter, they can be broadly classified into conditions affecting different parts of the human body. For example, codes starting with the letter G are related to conditions of the central nervous system. 

In order to obtain a feature representation of an investigator, we first aggregate all the insurance claims where the investigator is the rendering provider for the claim. We then aggregate just the first letters of the ICD-10 codes associated with these claims and form a 26 length vector with each entry equal to the number of times that letter appears in the investigator's treatment history. 

\textbf{Patient enrollment in past trials}

For each clinical trial, IQVIA's database contains a list of investigators who have enrolled in the trial in the past. For each of these investigators, we also have data regarding the investigator performance in the form of the number of patients enrolled. In this paper, we use the number of enrolled patients as a trial-specific score for the investigator. Past patient enrollment data reflects the patient access available to the investigator in general. Hence, it is a strong indicator of future enrollment performance. 

In our database, different trials have a different number of investigators with enrollment information available. For our experiments, we select a fixed number $M$ of these investigators for each trial. If the number of available investigators is less than $M$, we use negative sampling and select the remaining number of investigators from the general pool of investigators and set their past enrollment as zero.

\textbf{Patient group membership data}

The main goal of our proposed framework is to recommend investigators who have access to a diverse patient cohort, where diversity is defined in terms of the representation of different patient groups. In this paper, we use ethnicity based group definitions. To obtain the patient ethnic composition for each investigator, we use the investigator's zipcode and US census data available publicly at \url{https://statisticalatlas.com}. 

The above procedure results in a table with a list of zip codes and the corresponding ethnic composition. We assume that the same distribution can be used to model the ethnic composition of the patients that the investigator has access to.

\subsection{Model training and validation}

In our data, we have a total of 4264 clinical trials and a cumulative of 31836 investigators. In our setup, each clinical trial and the set of investigators to be ranked, together form a single data point.
We first divide the clinical trials according to their trial start date. We use the trials starting before Jan 01 2020 as our training and validation data. We use trials with the start date after the above date as test data. In particular, we have 3480 training trials, 304 validation trials and 480 test trials. 

 Further, ranking all of the 31,836 investigators for each trial is unnecessary and inefficient. In a real-world setting, it is common to obtain a list of potential investigators who are relevant to the clinical trial under consideration. In our experiments, we take the following approach: from the past enrollment data, we collect $M$ investigators with the highest patient enrollment. Our goal is then to obtain the Top-$K$ investigators from the list of $M$ investigators. In our experiments, we use $M=20, 100$ and  $K=10$.

% We use the same approach as above to validate/test our model. Given a validation/ test sample, we compare the model's list of Top-$K$ investigators to the set of investigators with the most patients enrolled 

\subsubsection{Baseline methods}
\label{subsubsec:baseline}

\textbf{Binary classification with cross entropy loss (BC):}
We compare our method with a binary classification based ranking method. For binary classification, we label the Top-$K$ investigators in the training data as 1 and the others as 0 and then use a multi-label classification setup. In particular, we use a sigmoid activation at the outputs of the final layer and then use a binary cross-entropy loss on each output node as described below:
\begin{equation}
    \mathcal{L}_{BCE} = -\sum_{m =1}^M y_m \cdot \log(h_\theta(m)) + (1-y_m) \cdot \log (1 - h_\theta(m)). 
\end{equation}

This baseline methods help us compare our method with a class of ranking methods that frame ranking as a binary classification problem  \cite{ rankNet, joachims2006training}. 

\textbf{Deterministic ranking based on score regression (regress.):}
Further, we also compare our method to a deterministic ranking method. We train a neural network based model to perform regression on the patient enrollment vector $\ve[\vt]$. To obtain the Top-$K$ ranking, we select the sites with the $K$-highest scores. This is equivalent to using a deterministic policy to generate a ranking where the policy just chooses the sites with $k$ highest scores.  The regression based model helps us compare our method to the class of existing methods that use score regression as the core method \cite{Listnet,ye2009stochastic}.

\textbf{Policy gradient with one-sided fairness loss (PG-OS):} Finally, we use the one-sided fairness loss proposed in \cite{singh2019ranking} as a baseline for fair ranking. We note that our method coincides when no diversity constraints are imposed. We also note once again that the one-sided fairness loss is not suitable under the multi-group membership setting. However, we present empirical comparisons with our method.

\subsubsection{Dataset statistics}

In this section, we provide insight into various attributes of our dataset. In particular, we provide details about the therapeutic areas targeted by the set of clinical trials in our dataset, the number of patients enrolled, and the number of investigators who undertook the study.

\begin{figure}
    \centering
    \includegraphics[scale=0.5]{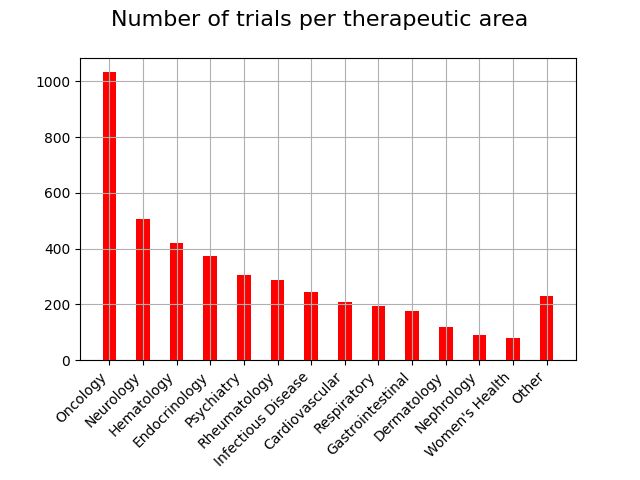}
    \caption{In our dataset, we have a total of 4264 clinical trials. These trials address various conditions that can be categorized into therapeutic areas. This bar plot shows the distribution of the trials into these categories.}
    \label{fig:TA_dist}
\end{figure}

Each clinical trial seeks to address a particular condition. These conditions can be categorized into different therapeutic areas. These therapeutic areas are one of the main factors influencing site selection and investigator selection for clinical trials. In our proposed approach, the therapeutic area is used as a feature to characterize clinical trials. In Figure \ref{fig:TA_dist}, we show the distribution of the clinical trials in our dataset across the standard therapeutic areas.

% \begin{table*}[t]
% \begin{center}
% \caption{Distribution of number of patients across the trials.}
% \label{tab:pat_per_trial}
%     % \resizebox{\textwidth}{!}{
%     \begin{tabular}{c|c}
%     \toprule
%     Number of patients & Number of trials \\
%     \midrule
%     Less than 50 & 2642\\
%     50-100 & 600 \\
%     100-500 & 894 \\
%     500-1000& 72\\
%     1000-5000 & 51\\
%     More than 5000 & 5\\
%     \bottomrule
%     \end{tabular}
% \end{center}
% \end{table*}

% \begin{table*}[t]
% \begin{center}
% \caption{Distribution of number of investigators across the trials.}
% \label{tab:inv_per_trial}
%     % \resizebox{\textwidth}{!}{
%     \begin{tabular}{c|c}
%     \toprule
%     Number of Investigators & Number of trials \\
%     \midrule
%     Less than 10 & 1502\\
%     10-50 & 1784 \\
%     50-100 & 567 \\
%     100-500& 393\\
%     More than 500 & 18\\
%     \bottomrule
%     \end{tabular}
% \end{center}
% \end{table*}

\begin{table*}[t]
\begin{center}
\caption{Distribution of trial phase across the trials.}
\label{tab:phase_per_trial}
    % \resizebox{\textwidth}{!}{
    \begin{tabular}{c|c}
    \toprule
    Phase & Number of trials \\
    \midrule
   Phase 1 & 863\\
   Phase 2 & 1318 \\
   Phase 3 & 1773 \\
   Phase 4 & 300\\
    \bottomrule
    \end{tabular}
\end{center}
\end{table*}

\section{Experimental results}

With the dataset and the framework as described above, we evaluate our model on multiple fronts. We first evaluate if our model is able to choose the best set of investigators based on just the past enrollment information. Then, we evaluate if adding the entropy based diversity constraints results in an improved and more equitable group distribution among the recruited patient cohort. To this end, we first show how the group representation changes on average across the clinical trial in our test dataset. Then, we present case studies on individual clinical trials. Finally, we analyze the performance of our model across different theoretical areas and trial phases and present our findings.

\subsection{Evaluation metrics}
\label{subsec:eval_metrics}
We use the following metrics to measure  the utility and fairness of our model.

\begin{enumerate}
    \item \textbf{Relative error:} To measure the ability to rank, we compare the number of patients expected to be enrolled by the set of Top-$K$ investigators as ranked by the model to the maximum possible number of patients that can be enrolled by any set of $K$ investigators from the list of $M$ investigators. Let $\ve_s(\vt)$ be the vector of sorted enrollment rates for trial $\vt$ in the decreasing order. We define the maximum patient enrollment and expected patient enrollment as $\vt$ as:
\begin{align}
    \Delta_{\max}(\vt) & = \sum_{i=1}^K \ve_s(\vt)[i] \\
    \Delta_\pi(\vt) & = \E_{r \sim \pi} \sum_{j=1}^K \ve(\vt)[r[j]].
\end{align} Finally, we compute the relative error for trial $\vt$ as 

\begin{equation}
    \text{relative error} = \Delta(\vt) = \frac{ \Delta_{\max}(\vt) -  \Delta_{\pi}(\vt)}{ \Delta_{\max}(\vt)}
\end{equation}
    
\item \textbf{Recall:} We define recall as the fraction of top-K items selected by our model:
\begin{equation}
    \text{Recall:} \frac{|\text{Number of top-K sites selected}|}{
    |\text{Total number of top-K sites}|}
\end{equation}
Note that the total number of top-K sites is equal to a minimum of the number of sites enrolled in the trial and $K=$10.

\item \textbf{NDCG:} We also compute the top-$K$ Normalized Discounted Cumulative Gain (NDCG), which is a standard ranking metric. It is defined as 
\begin{equation}
    \text{NDCG} = \sum_{k=1}^{K} \frac{2^{\ve[i]} - 1}{\log_2(i+1)} / \sum_{k=1}^{K} \frac{2^{\ve_s[i]} - 1}{\log_2(i+1)}
\end{equation}
    
    \item \textbf{Entropy:} We measure the diversity of a particular ranking by the entropy of the average group representation over the Top-$K$ set of sites. A higher entropy denotes a more uniform group representation. Along with entropy, we also present the resulting group representations to demonstrate that an increased entropy indeed results in a more diverse patient cohort.

\end{enumerate}

For all experiments, we use a neural network with 2 hidden layers of size 1024 and 256 and ReLU activations. The output layer has a Softmax activation.

\subsection{Trial site matching without fairness - optimizing for utility}

Our first set of experiments seek to evaluate if the framework of policy-based ranking described above can choose the best set of $K$ investigators from a larger list of size $M$.  Note that in this case, our method and PG-OS coincide (modulo the  utility function). To this end, we set the regularization coefficient $\lambda$ to 0 and train the model with the set of clinical trials in the training set. For each trial in the test set, we select a list of $M$ investigators from our database who have previously enrolled in that clinical trial and our model ranks the Top-$K$ from this list. For trials with less than $M$ active investigators, we include a randomized set from the pool of investigators. 

\begin{figure}
    % \centering
     \begin{subfigure}[b]{0.5\textwidth}
         \centering
         \includegraphics[scale=0.35]{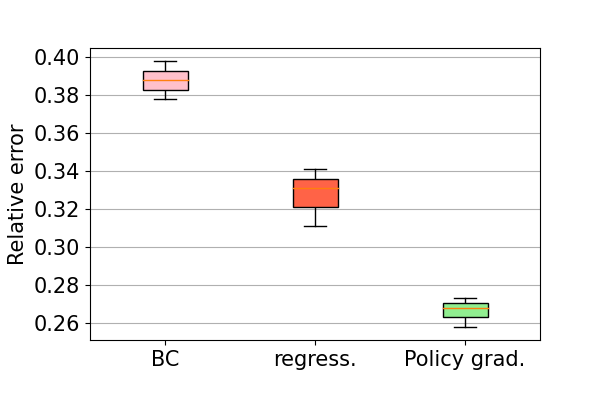}
         \caption{$M=20$, $K=10$}
         \label{fig:M20_perf}
     \end{subfigure}
     \hfill
     \begin{subfigure}[b]{0.5\textwidth}
         \centering
         \includegraphics[scale=0.35]{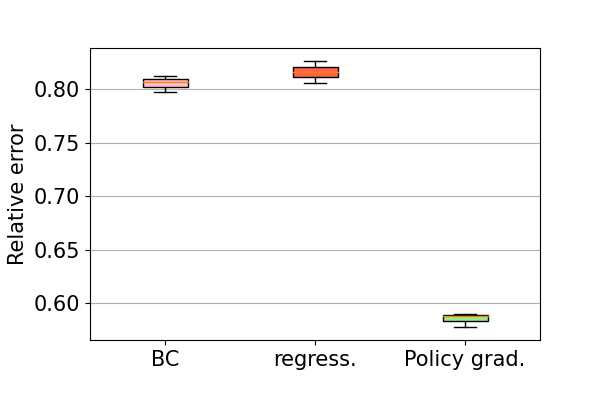}
         \caption{$M=100$, $K=10$}
         \label{fig:M100_perf}
     \end{subfigure}
 \caption{Relative error in expected patient enrollment across the three baselines and our proposed method. One the left, $M=20$ and for the figure on the right, $M=100$. BC stands for binary classification based baseline, `regress.' stands for regression based ranking and `Policy grad.' is our method. Note that our method maintains a low error even when the method is scaled up to $M=100$. }
        \label{fig:utility_perf}
\end{figure}

We experiment with two values of $M$, $20$, and $100$. Note that choosing top-10 from a list of $100$ investigators is a harder task than when $M=20$. We present the results of this experiment in Figures \ref{fig:M20_perf} and \ref{fig:M100_perf}. In both cases, we can observe that ranking based on binary classification and score-regression perform poorly. The performance degrades further when $M=100$. This is mainly caused by the deterministic nature of the ranking policy used by these methods. For multi-modal and high dimensional data such as ours, simple baselines may not have the capacity to perform well. These results serve as evidence that our ranking algorithm without fairness constraints (which coincides with \cite{singh2019ranking}) performs well on the complex dataset under consideration. In Table \ref{tab:ranking_metrics_results}, we present quantitative performace in terms of standard ranking metrics described in Subsection \ref{subsec:eval_metrics}.  In the sequel, we present the fairness performance of our proposed entropy-based mechanism.

\begin{table*}[h]
    \centering
        \caption{Comparison of ranking metrics across baseline methods and our proposed method. }
    \label{tab:ranking_metrics_results}
    \begin{tabular}{l|ccc|ccc}
    \toprule
    Method & \multicolumn{6}{c}{Ranking metrics without diversity constraints}\\
    \midrule
    & \multicolumn{3}{c}{M = 20} & \multicolumn{3}{c}{M = 100}\\
    \midrule 
    & Rel. err & Recall & NDCG & Rel. err & Recall & NDCG  \\
    \midrule
    BC & 0.39 & 0.56  & 0.40  & 0.8 & 0.17 & 0.13 \\
    regress. & 0.33 &  0.60 & 0.46 & 0.70 & 0.25  & 0.20  \\
    \textbf{PG (ours)} &  \textbf{0.27} & \textbf{0.68}  & \textbf{0.53} & \textbf{0.58} &\textbf{ 0.38 }& \textbf{0.29}\\
    % PG-entropy & 0.27 & && &&  \\
    \bottomrule
    \end{tabular}
\end{table*}

% {\color{red} same ylim for both figures. Or use a single plot with color code and plot methods side-by-side. OR, parallel coordinate plot.}

\subsection{Diversity constrained policy learning}

Our next goal is to study whether our framework can learn to simultaneously optimize for both the number of patients expected to be enrolled and reduce representation disparity amongst the different groups. We explore this by setting $\lambda > 0$. We tune this parameter based on the validation dataset. We study two competing goals: i) improve in group representation ii) minimize the reduction in the model utility due to fairness constraints.

\subsubsection{Average model performance}

\begin{figure}
    \centering
    \begin{subfigure}[b]{0.48\textwidth}
         \centering
         \includegraphics[scale=0.35]{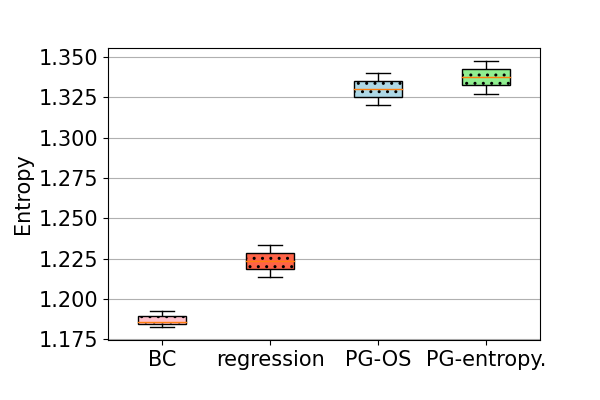}
         \caption{$M=20$, $K=10$}
         \label{fig:M20_entropy}
     \end{subfigure}
     \hfill
    %  \begin{subfigure}[b]{0.48\textwidth}
        %  \centering
        %  \includegraphics[scale=0.35]{figures/performance_plots/entropy_M100_w_OS.png}    
        %  \caption{$M=100$, $K=10$}
        %  \label{fig:M100_entropy}
    %  \end{subfigure}
    \caption{The entropy of the average group representation of the Top-$K$ sites is chosen by the ranking algorithm. Note that our method achieves the highest entropy, resulting in a diverse patient cohort.}
    \label{fig:entropy}
\end{figure}

In our first set of results, we study the average change in the group representation and the average change in the expected patient enrollment. For each trial in the test set, we sample 20 rankings based on the model's score and the ranking policy. We compute the average entropy across the ranking samples and then average it over all the trials in the test dataset. In Figure \ref{fig:entropy}, we illustrate the average entropy of group representation across clinical trials in the test dataset. As can be seen, our proposed approach achieves the highest entropy compared to the baseline methods. 

\begin{figure}
    \centering
    \includegraphics[width=1\textwidth, height=2.5in]{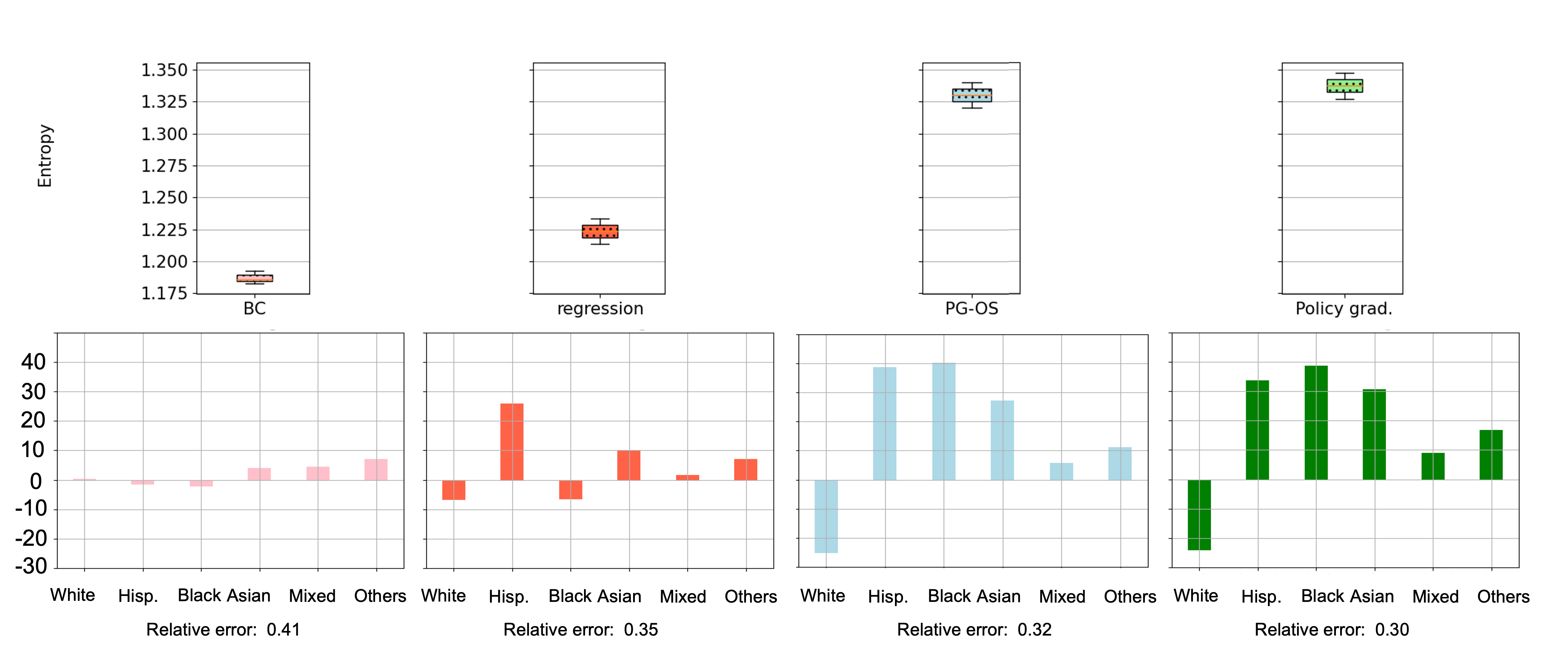}
    \caption{Average relative change in the group representation across the clinical trials in the test set.}
    \label{fig:pop_rel_change}
\end{figure}

We then investigate how the gain in entropy translates to group representations. For each ranking we compute the average group representation of the Top-$K$ set of investigators and then average this over the 20 rank samples. We compute the expected group representation of the model, $\vn^{*}$ as:
\begin{align}
    \vn^{*}[\ell] & = \E_{r \sim \pi} \frac{1}{K} \sum_{k=1}^K \mP[r[k], \ell], \ \ {\ell \in \{1, \ \cdots, L \}}\\
\end{align} where the expectation is computed using the samples of the ranking. We then compute the average group representation across all the trials in the test dataset. In Figure \ref{fig:pop_rel_change}, we show the relative change in the group representation with and without diversity constraints for each of the baseline methods and our proposed method. Note that the relative change in representation is defined as 
\begin{equation}
    \mathrm{rel. change} = \frac{\% \text{with div. constraints} - \text{\% without div. constraints}}{\text{\% without div. constraints}}.
\end{equation}
The Binary classification and regression based methods do not lead to any meaningful changes in the group representation. They also have a higher reduction in the expected patient enrollment rate. In the third baseline, we implement the one-sided fairness loss proposed by \cite{singh2019ranking}. Although this method achieves good empirical performance, is is sub-optimal in encouraging diversity in the case of multi-group membership as explained in Section \ref{subsec:fair_setup}. As can be seen, our proposed entropy based method outperforms all of the other baseline methods.

We study two cases, with $M=20$ and $M=100$. In Tables \ref{tab:ranking_fairness_metrics_results_M20} and \ref{tab:ranking_fairness_metrics_results_M100}, we present quantitative performance in terms of standard ranking metrics described in Subsection \ref{subsec:eval_metrics} for both cases. From the results, it can be seen that the entropy-based fairness metric outperforms other baseline methods across all the metrics.


\begin{table*}[h]
    \centering
        \caption{Comparison of ranking and diversity metrics across baseline methods and our proposed method. }
    \label{tab:ranking_fairness_metrics_results_M20} \begin{tabular}{l|cccc}
    \toprule
    Method & \multicolumn{4}{c}{Metrics with diversity constraints, M = 20}\\
    \midrule 
    & Rel. err & Recall & NDCG & Entropy \\
    \midrule

    BC + negative entropy & 0.41& 0.53& 0.39 & 1.19    \\
    regress. + negative entropy & 0.35 & 0.58 & 0.45 & 1.22  \\
    PG-OS (\cite{singh2019ranking}) & 0.32 & \textbf{0.62} & 0.47& 1.33 \\
    \textbf{PG-entropy (ours)} & \textbf{0.30 }& \textbf{0.62} & \textbf{0.49}&\textbf{ 1.34}\\
    \bottomrule
    \end{tabular}
\end{table*}

\begin{table*}[h!]
    \centering
        \caption{Comparison of ranking and diversity metrics across baseline methods and our proposed method. }
    \label{tab:ranking_fairness_metrics_results_M100} \begin{tabular}{l|cccc}
    \toprule
    Method & \multicolumn{4}{c}{Metrics with diversity constraints, M = 100}\\
    \midrule 
    & Rel. err & Recall & NDCG & Entropy \\
    \midrule
    BC & 0.86 & 0.12 & 0.10 & 1.15   \\
    regress. & 0.72  & 0.23 & 0.19 & 1.16 \\
    PG-OS & 0.69  & 0.27  & 0.21 & \textbf{1.35}\\
    \textbf{PG-entropy (ours)} &  \textbf{0.60}& \textbf{0.36} &\textbf{0.29} & 1.30\\
    \bottomrule
    \end{tabular}
\end{table*}

\subsection{Case studies on individual trials}
 We now present some case studies on individual clinical trials that are part of our test dataset to provide more insight into the workings of our algorithm. We find that for all of the clinical trials in our test set, the group representation improves. We find that the degree of improvement depends on the available list of sites the algorithm can choose from. For ease of presentation, we provide results with $M=20$ and $K=10$.

\textbf{Case study 1: Trial NCT ID  NCT04369053}
Prevention of Colorectal Cancer Through Multiomics Blood Testing (\url{https://www.clinicaltrials.gov/ct2/show/NCT04369053}). For this trial, the locations of the investigators that were provided to the algorithm during inference is presented in Table \ref{tab:T1_input}. The listed site locations are in the decreasing order of expected patient enrollment. 
For this trial, our model chooses the sites highlighted in bold font in Table \ref{tab:T1_input} when we set $\lambda=0$. As can be seen, our model is able to choose 7 out of the top-10 correctly.

\begin{figure}
    \centering
    \includegraphics[scale=0.5]{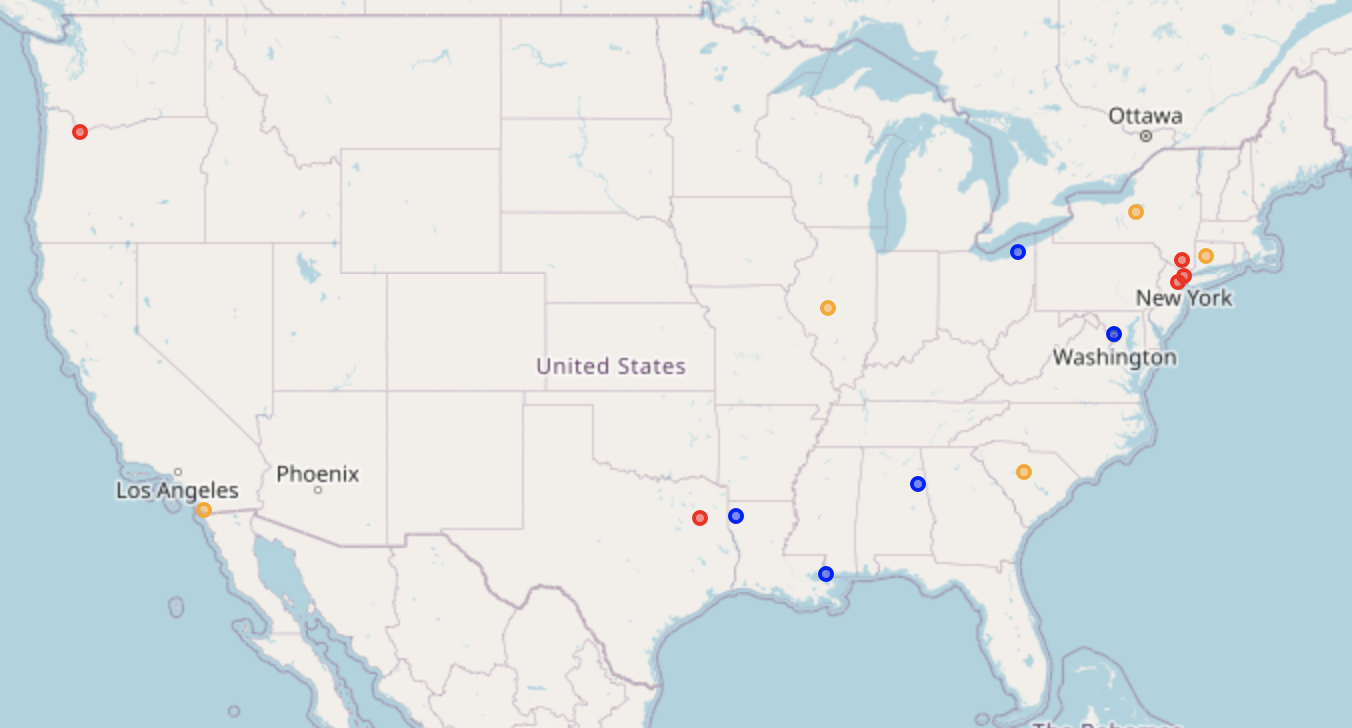}
    \caption{Geographic locations of the trial sites chosen by our model with and without the diversity constraints. The sites marked in orange are common to both lists. When diversity constraints are added, the model drops the sites marked in red and adds those marked in blue.}
    \label{fig:T1_map}
\end{figure}

We then use a model trained with $\lambda=4$ to encourage the recruitment of a more diverse patient cohort. As can be seen in Figure, this results in a much more diverse patient cohort.  Let us also explore \textbf{how} such a change is brought about in group representation. Let us consider the set of investigators that were \textbf{dropped} from the Top-$K$ list and the set of investigators that were \textbf{added} to the Top-$K$ list based on diversity constraints. We list the location of these investigators in Table \ref{tab:T1_dropped} and \ref{tab:T1_added}. We also visualize this on the map of the USA in Figure \ref{fig:T1_map}. As can be seen, the model chooses investigators from locations with more diverse demographics. At the same time, the model also retains the highly ranked sites in order to preserve expected enrollment.

\begin{table*}[h]
\begin{center}
\caption{Case study 1: Trial NCT04369053 The list of 20 investigators provided as input to the algorithm, along with their associated race and ethnicity distribution. The highlighted investigators are the ones chosen by our algorithm when $\lambda=0$. Note that our algorithm identifies 7 out of the top-10 correctly.}
\label{tab:T1_input}
\begin{tabular}{l|ccccccc}
\toprule
           Investigators (Locations) &  White &  Hispanic &  Black &  Asian &  Mixed &  Others & Num. pat. \\
\midrule
     \textbf{Slidell, LA} &   77.0 &       4.3 &   15.3 &    1.2 &    1.7 &     0.6 &                183.0 \\
    \textbf{Syracuse, NY} &   58.2 &       6.2 &   19.3 &   11.5 &    3.4 &     1.4 &                169.0 \\
 New Windsor, NY &   60.1 &      21.7 &   10.9 &    4.5 &    2.4 &     0.4 &                140.0 \\
      \textbf{Mentor, OH} &   95.6 &       1.2 &    0.7 &    1.4 &    1.0 &     0.1 &                134.0 \\
    \textbf{Columbia, SC} &   12.5 &       3.0 &   82.2 &    0.6 &    1.5 &     0.1 &                130.0 \\
 \textbf{Chevy Chase, MD} &   78.2 &       7.5 &    4.1 &    6.7 &    2.9 &     0.7 &                130.0 \\
  \textbf{Shreveport, LA} &   78.6 &       3.7 &   11.4 &    3.5 &    2.4 &     0.5 &                114.0 \\
     Atlanta, GA &   62.6 &      21.9 &    9.1 &    4.4 &    2.0 &     0.1 &                108.0 \\
 \textbf{Springfield, IL} &   77.0 &       1.7 &   16.8 &    1.1 &    2.8 &     0.6 &                 88.0 \\
     Teaneck, NJ &   44.7 &      15.5 &   29.3 &    8.5 &    1.4 &     0.6 &                 76.0 \\
      Topeka, KS &   75.2 &      12.1 &    6.4 &    0.5 &    4.8 &     1.0 &                 64.0 \\
   \textbf{Waterbury, CT} &   57.0 &      21.3 &   14.9 &    1.9 &    3.5 &     1.4 &                 60.0 \\
       Union, NJ &   42.3 &      17.3 &   26.3 &   10.9 &    0.9 &     2.4 &                 60.0 \\
    Portland, OR &   61.2 &      12.2 &    8.4 &   11.9 &    4.3 &     2.0 &                 59.0 \\
   \textbf{San Diego, CA} &    9.4 &      41.9 &   19.5 &   24.9 &    3.6 &     0.7 &                 55.0 \\
        Bend, OR &   87.4 &       7.9 &    0.5 &    1.7 &    2.2 &     0.3 &                 54.0 \\
     Wichita, KS &   73.1 &       5.6 &   12.6 &    5.8 &    2.8 &     0.1 &                 43.0 \\
  Union City, TN &   78.3 &       4.5 &   14.7 &    0.1 &    1.9 &     0.3 &                 39.0 \\
       Tyler, TX &   10.4 &      49.1 &   39.3 &    0.2 &    0.6 &     0.5 &                 32.0 \\
    \textbf{Anniston, AL} &   69.1 &       3.2 &   24.7 &    1.3 &    1.6 &     0.1 &                 30.0 \\
\bottomrule
\end{tabular}
\end{center}
\end{table*}

\begin{table*}[h]
\begin{center}
\caption{The list of sites dropped from the chosen list when $\lambda=4$. Note that the dropped sites have a very homogeneous patient population. Our algorithm chooses other sites with a more diverse population in lieu of these sites.}
\label{tab:T1_dropped}
\begin{tabular}{l|ccccccc}
\toprule
            Investigators(Locations) &  White &  Hispanic &  Black &  Asian &  Mixed &  Others \\
\midrule
     Slidell, LA &   77.0 &       4.3 &   15.3 &    1.2 &    1.7 &     0.6\\
      Mentor, OH &   95.6 &       1.2 &    0.7 &    1.4 &    1.0 &     0.1  \\
 Chevy Chase, MD &   78.2 &       7.5 &    4.1 &    6.7 &    2.9 &     0.7  \\
  Shreveport, LA &   78.6 &       3.7 &   11.4 &    3.5 &    2.4 &     0.5   \\
    Anniston, AL &   69.1 &       3.2 &   24.7 &    1.3 &    1.6 &     0.1 \\
\bottomrule
\end{tabular}
\end{center}
\end{table*}

\begin{table*}[h]
\begin{center}
\caption{The list of sites added  when $\lambda=4$. Note that the added sites have a higher representation of different communities (compared to the dropped sites in Table \ref{tab:T1_dropped}), leading to a more diverse patient population.}
\label{tab:T1_added}
\begin{tabular}{lrrrrrrr}
\toprule
            Investigators (Locations) &  White &  Hispanic &  Black &  Asian &  Mixed &  Others  \\
\midrule
 New Windsor, NY &   60.1 &      21.7 &   10.9 &    4.5 &    2.4 &     0.4\\
     Teaneck, NJ &   44.7 &      15.5 &   29.3 &    8.5 &    1.4 &     0.6 \\
       Union, NJ &   42.3 &      17.3 &   26.3 &   10.9 &    0.9 &     2.4  \\
    Portland, OR &   61.2 &      12.2 &    8.4 &   11.9 &    4.3 &     2.0  \\
       Tyler, TX &   10.4 &      49.1 &   39.3 &    0.2 &    0.6 &     0.5 \\
\bottomrule
\end{tabular}
\end{center}
\end{table*}

\begin{figure}[h!]
    % \centering
     \begin{subfigure}[b]{0.45\textwidth}
         \centering
         \includegraphics[scale=0.3]{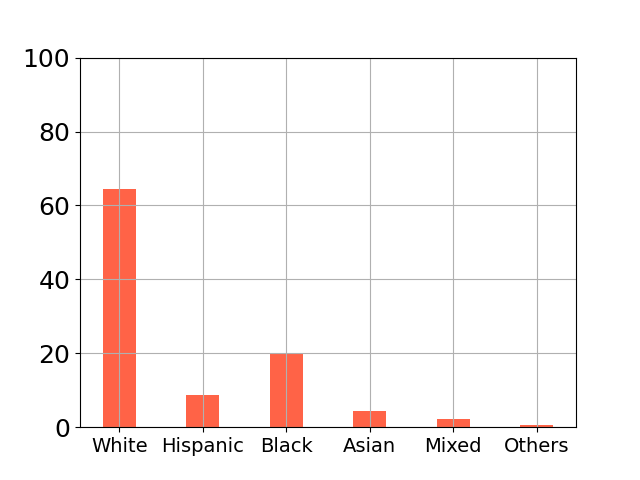}
         \caption{}
         \label{fig:T1_base_rep}
     \end{subfigure}
     \hfill
     \begin{subfigure}[b]{0.45\textwidth}
         \centering
         \includegraphics[scale=0.3]{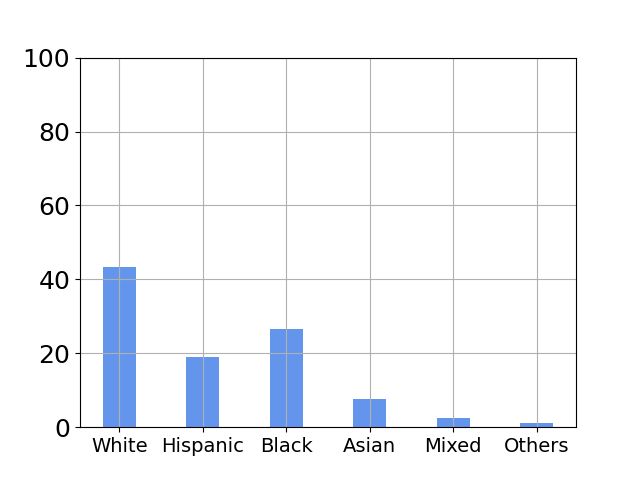}
         \caption{}
         \label{fig:T1_model_rep}
     \end{subfigure}
        \caption{Expected patient group representations for  NCT04369053 with and without diversity constraints. Note that the representation is more equitable in the latter case. }
        \label{fig:T1_perf}
\end{figure}

\textbf{Case study 2: Trial NCT ID NCT04625725}

Placebo-controlled Study of AZD7442 for Pre-exposure Prophylaxis of COVID-19 in Adult (\url{https://www.clinicaltrials.gov/ct2/show/NCT04625725}). For this trial, the locations of the investigators that were provided to the algorithm during inference are presented in Table \ref{tab:T2_input}. The listed site locations are in the decreasing order of expected patient enrollment. 
For this trial, our model chooses the sites highlighted in bold font in Table \ref{tab:T2_input} when we set $\lambda=0$. As can be seen, our model is able to choose 8 out of the top-10 correctly.

\begin{figure}[h]
    \centering
    \includegraphics[scale=0.5]{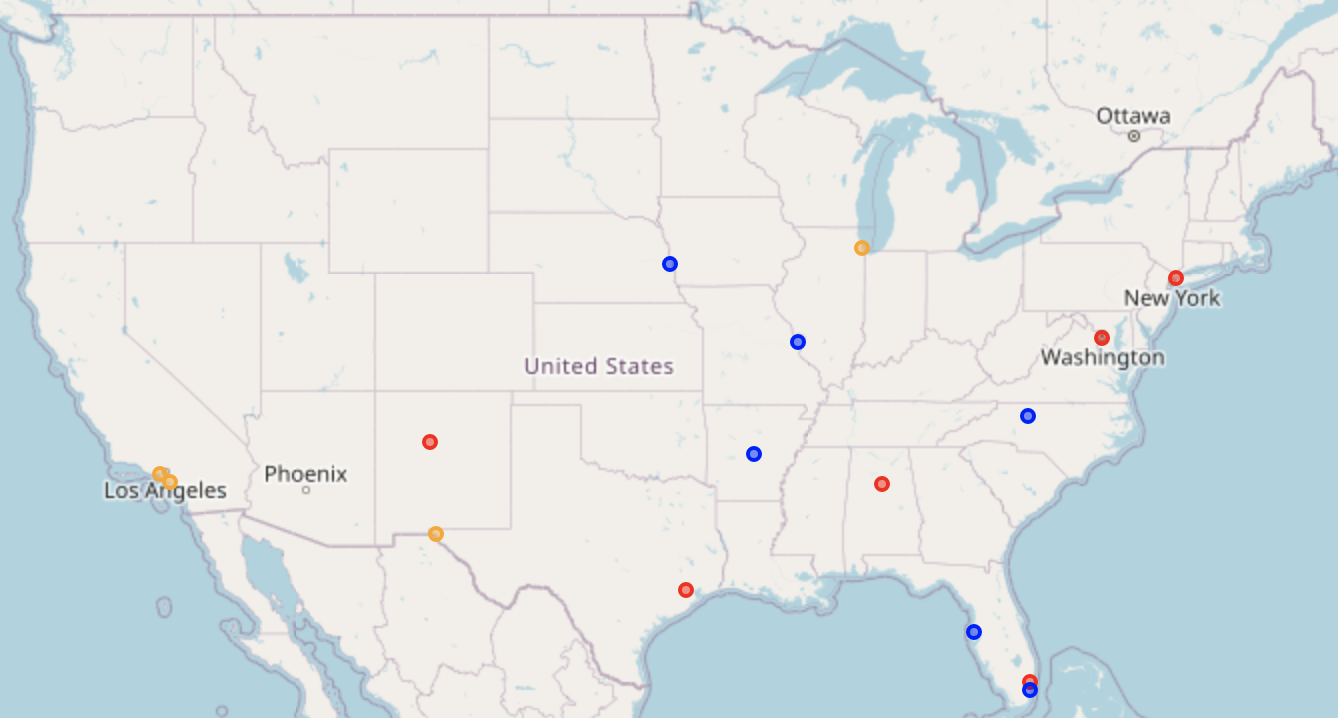}
    \caption{Geographic locations of the trial sites chosen by our model with and without the diversity constraints for trial NCT04625725. The sites marked in orange are common to both the lists. When diversity constraints are added, the model drops the sites marked in red and adds those marked in blue. }
    \label{fig:T2_map}
\end{figure}

We then use a model trained with $\lambda=4$ to encourage the recruitment of a more diverse patient cohort. As can be seen in Figure, this results in a much more diverse patient cohort.  Let us also explore \textbf{how} such a change is brought about in group representation. Let us consider the set of investigators that were \textbf{dropped} from the Top-$K$ list and the set of investigators that were \textbf{added} to the Top-$K$ list based on diversity constraints. We list the location of these investigators in Table \ref{tab:T2_dropped} and \ref{tab:T2_added}. We also visualize this on the map of the USA in Figure \ref{fig:T2_map}. As can be seen, the model chooses investigators from locations with a more diverse demographics. At the same time, the model also retains the highly ranked sites in order to preserve expected enrollment.

\begin{table*}[h]
\begin{center}
\caption{Case study 2: Trial NCT04625725 The list of 20 investigators provided as input to the algorithm, along with their associated race and ethnicity distribution. The highlighted investigators are the ones chosen by our algorithm when $\lambda=0$. Note that our algorithm identifies \textbf{8 out of the top-10} correctly.}
\label{tab:T2_input}
\begin{tabular}{lccccccc}
\toprule
Investigators(Locations) &  White &  Hispanic &  Black &  Asian &  Mixed &  Others &  Expected Enrollment \\
\midrule
         \textbf{Little Rock, AR} &   64.5 &       3.1 &   25.8 &    3.7 &    2.2 &     0.8 &                216.0 \\
             \textbf{Chicago, IL} &   22.9 &       9.0 &   25.5 &   40.7 &    1.8 &     0.2 &                137.0 \\
         \textbf{Saint Louis, MO} &   77.1 &       2.9 &    6.8 &   11.0 &    1.4 &     0.8 &                 61.0 \\
       \textbf{Winston-Salem, NC} &   58.5 &      14.0 &   21.8 &    4.1 &    1.6 &     0.1 &                 60.0 \\
          Birmingham, AL &   22.8 &       4.6 &   69.6 &    0.3 &    2.7 &     0.1 &                 60.0 \\
         \textbf{Palm Harbor, FL} &   87.0 &       4.9 &    3.5 &    2.1 &    2.0 &     0.4 &                 39.0 \\
               \textbf{Miami, FL }&    3.8 &      87.0 &    7.6 &    1.4 &    0.1 &     0.0 &                 37.0 \\
         Albuquerque, NM &   51.7 &      32.7 &    4.1 &    3.2 &    2.2 &     6.1 &                 35.0 \\
         \textbf{Los Angeles, CA} &   53.2 &      18.4 &    2.9 &   20.4 &    4.9 &     0.2 &                 26.0 \\
             \textbf{El Paso, TX} &   13.9 &      81.2 &    4.2 &    0.3 &    0.2 &     0.2 &                 22.0 \\
     Fort Lauderdale, FL &    9.8 &       7.2 &   78.8 &    1.7 &    1.7 &     0.8 &                 21.0 \\
          Evansville, IN &   73.8 &       6.2 &   14.2 &    0.4 &    4.5 &     0.9 &                 21.0 \\
               Bronx, NY &   35.6 &      42.8 &    8.9 &   11.1 &    1.2 &     0.4 &                 18.0 \\
         \textbf{Westminster, CA} &   24.9 &      23.7 &    0.8 &   47.8 &    2.2 &     0.5 &                 16.0 \\
               \textbf{Omaha, NE} &   71.8 &       4.4 &   17.7 &    2.6 &    3.1 &     0.5 &                 15.0 \\
           Ridgewood, NY &   47.2 &      42.7 &    2.3 &    6.7 &    0.8 &     0.3 &                 12.0 \\
              Austin, TX &   70.5 &      17.0 &    2.4 &    6.4 &    2.6 &     1.1 &                 11.0 \\
          Chesapeake, VA &   55.2 &       6.2 &   29.8 &    5.4 &    3.0 &     0.4 &                 11.0 \\
          Alexandria, VA &   41.7 &      16.2 &   28.4 &    9.2 &    4.2 &     0.3 &                  9.0 \\
             Houston, TX &   29.8 &       8.0 &   37.3 &   22.6 &    1.8 &     0.5 &                  8.0 \\
\bottomrule
\end{tabular}
\end{center}
\end{table*}

\begin{table*}[h]
\begin{center}
\caption{The list of sites dropped from the chosen list when $\lambda=4$. Note that the dropped sites have a high percentage of white population, leading to a homogeneous patient population.}
\label{tab:T2_dropped}
\begin{tabular}{lccccccc}
\toprule
Investigators(Locations) &  White &  Hispanic &  Black &  Asian &  Mixed &  Others &  Expected Enrollment \\
\midrule
         Little Rock, AR &   64.5 &       3.1 &   25.8 &    3.7 &    2.2 &     0.8 &                216.0 \\
         Saint Louis, MO &   77.1 &       2.9 &    6.8 &   11.0 &    1.4 &     0.8 &                 61.0 \\
       Winston-Salem, NC &   58.5 &      14.0 &   21.8 &    4.1 &    1.6 &     0.1 &                 60.0 \\
         Palm Harbor, FL &   87.0 &       4.9 &    3.5 &    2.1 &    2.0 &     0.4 &                 39.0 \\
               Miami, FL &    3.8 &      87.0 &    7.6 &    1.4 &    0.1 &     0.0 &                 37.0 \\
\bottomrule
\end{tabular}
\end{center}
\end{table*}

\begin{table*}[h!]
\begin{center}
\caption{The list of sites added  when $\lambda=4$. Note that the added sites have a higher representation of different communities (compared to the dropped sites in Table \ref{tab:T2_dropped}), leading to a more diverse patient population.}
\label{tab:T2_added}
\begin{tabular}{lccccccc}
\toprule
Investigators(Locations) &  White &  Hispanic &  Black &  Asian &  Mixed &  Others &  Expected Enrollment \\
\midrule
     Fort Lauderdale, FL &    9.8 &       7.2 &   78.8 &    1.7 &    1.7 &     0.8 &                 21.0 \\
               Bronx, NY &   35.6 &      42.8 &    8.9 &   11.1 &    1.2 &     0.4 &                 18.0 \\
         Westminster, CA &   24.9 &      23.7 &    0.8 &   47.8 &    2.2 &     0.5 &                 16.0 \\
          Alexandria, VA &   41.7 &      16.2 &   28.4 &    9.2 &    4.2 &     0.3 &                  9.0 \\
             Houston, TX &   29.8 &       8.0 &   37.3 &   22.6 &    1.8 &     0.5 &                  8.0 \\
\bottomrule
\end{tabular}
\end{center}
\end{table*}

\begin{figure}[h!]
    % \centering
     \begin{subfigure}[b]{0.45\textwidth}
         \centering
         \includegraphics[scale=0.3]{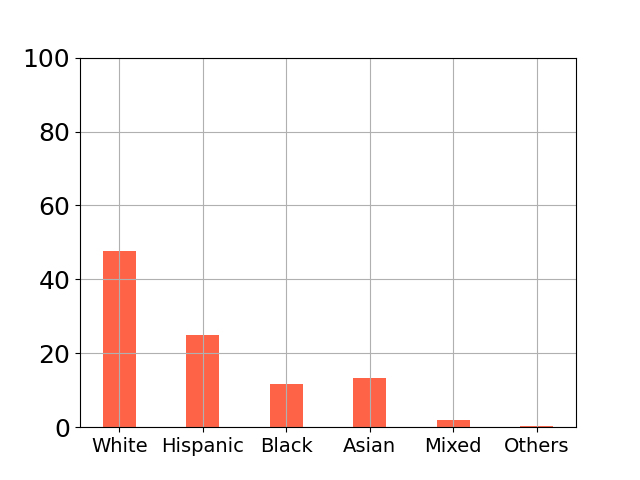}
         \caption{}
         \label{fig:T2_base_rep}
     \end{subfigure}
     \hfill
     \begin{subfigure}[b]{0.45\textwidth}
         \centering
         \includegraphics[scale=0.3]{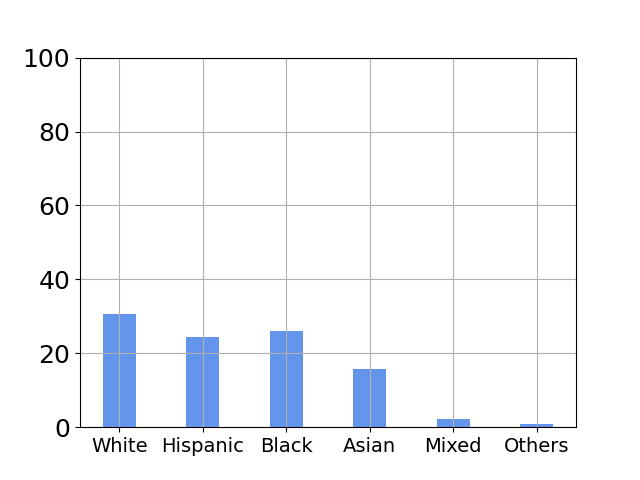}
         \caption{}
         \label{fig:T2_model_rep}
     \end{subfigure}
        \caption{Expected patient group representations for  NCT04625725 with and without diversity constraints. Note that the representation is more equitable in the latter case. }
        \label{fig:T2_perf}
\end{figure}

\subsection{Model performance across therapeutic areas and trial phases}

Clinical trials are characterized by many features, including the primary therapeutic area of the condition they are aimed at and the phase of the trial. Studying representation disparities across therapeutic areas can aid in better designing the site selection process. For instance, certain conditions and are more common in specific demographic groups. Our model can aid in selecting sites for such conditions so as to attain a more representative patient cohort.

In Table \ref{tab:TA_perf}, we show the relative change in the group representations for each therapeutic area. Note that in general, we expect a small negative relative change in the population of the white community (since the initial representation is large) and we expect a large positive relative change in the representation of the other communities. From Table \ref{tab:TA_perf}, this is indeed the general trend. We also note that there is a case where the representation of the 'Others' group also reduces. For this, we analyzed the expected change in the absolute patient volume based on the expected patient enrollment. We find that the absolute patient volume for this group is very small to begin with and the associated reduction in expected patient volume is also small. We highlight these details in Table \ref{tab:neg_TA_perf}. 

\begin{figure}[h!]
    \centering
    \includegraphics[scale=0.6]{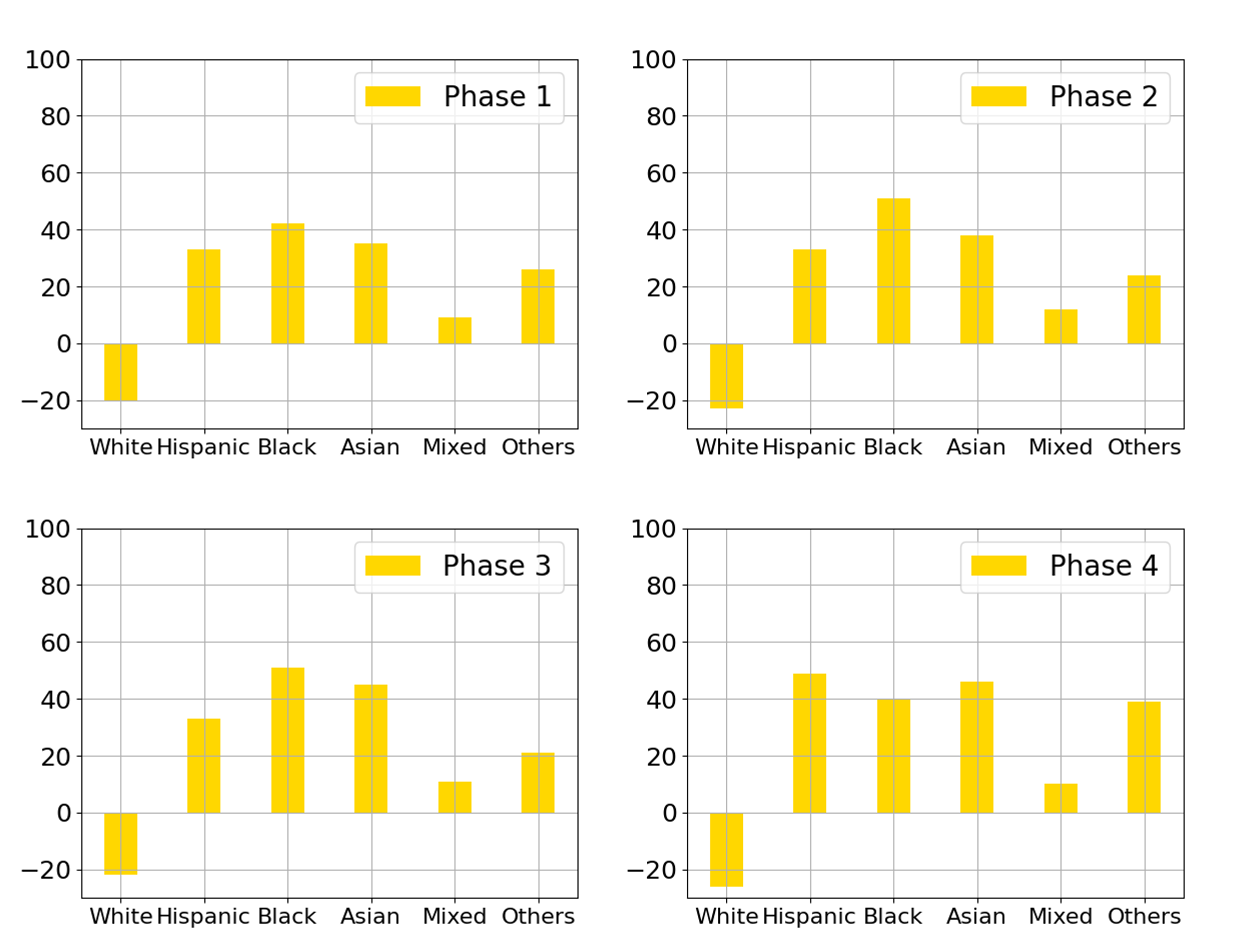}
    \caption{The relative change in the representation of the 6 racial groups across different trial phases.}
    \label{fig:phase_perf}
\end{figure}

On a similar note, we also break down the model performance across different trial phases.  We present the results in Figure \ref{fig:phase_perf}. Again, we observe a similar trend as before where the white community has a small negative relative change whereas the other communities have a significant positive relative change.

\section{Conclusion}

In this study, we consider the problem of aligning the study population and treatment population in the context of clinical trials. Our proposed method is able to learn the complex associations between patient enrollment in a clinical trial and the clinical investigator and their treatment history. Our proposed entropy-based fairness metric is able to handle the multi-group membership of the candidate investigators. We find that our algorithm is consistently able to improve the diversity of the enrolled patient cohort, while preserving the number of enrolled patients.
Our hope is that this study can lead to further exploration of developing trial-site identification that can account for additional factors such as stronger priors on patient demographics that are trial-specific, and the temporal nature of patient enrollment.

% \begin{table*}[h]
% \small
% \begin{center}
% \caption{The relative percentage change in the representation of the 6 racial groups across different therapeutic areas. {\color{red} add patient volume for each TA.} }
% \label{tab:TA_perf}
%     \begin{tabular}{l|cccccc}
%     \toprule
% Therapeutic area & White (\%) & Hispanic(\%) & Black(\%) & Asian(\%) & Mixed(\%) & Others(\%) \\    
% \midrule
% Ophthalmology &  -29& +128&  +100&  +99& -10& -4 \\
% Infectious Disease &  -15&  +11&  +94&  +15&  +12&  +43 \\
% Respiratory &  -11&  +38&  +64&  0&  +5&  +42 \\
% Dermatology &  -12&  +45&  +41&  +13&  +6&  +82 \\
% Nephrology &  -7&  +4&  +35&  +12&  +51&  +158 \\
% Rheumatology &  -7&  +13&  +30&  +11& -5& 0 \\
% Immunology &  -7&  +6&  +28&  +12& -7&  +11 \\
% Women's Health &  -7&  +15&  +31&  +1& -4& -2 \\
% Hepatology &  -10&  +22&  +34&  +3& -9& -11 \\
% Psychiatry &  -10&  +9&  +25&  +3&  +3&  +12 \\
% Gastrointestinal &  -6&  +12&  +20& 0& -1&  +7 \\
% Neurology &  -8&  +15&  +22&  +7&  +6&  +28 \\
% Oncology &  -8&  +19&  +16&  +3& -8&  +3 \\
% Endocrinology &  -6&  +4&  +18&  +37& 0&  +6 \\
% Cardiovascular &  -9&  +28&  +13&  +7&  +4&  +17 \\
% Hematology &  -7&  +29&  +6& 0& 0&  +21 \\
% Other &  -15&  +62&  +16&  +24& -25&  +1 \\
% \bottomrule
% \end{tabular}
% \end{center}
% \end{table*}

\begin{table*}[h]
\small
\begin{center}
\caption{The relative percentage change in the representation of the 6 racial groups across different therapeutic areas.}
\label{tab:TA_perf}
\begin{tabular}{l | cccccc}
\toprule
Therapeutic area &   White(\%) & Hispanic(\%) &  Black(\%) &   Asian(\%) &  Mixed(\%) & Others(\%) \\
\midrule
Allergy                        &  -21.16 &    34.26 &  89.64 &   33.55 &  10.86 &   7.54 \\
Neurology                      &  -24.10 &    40.48 &  58.25 &   33.50 &   7.25 &  28.08 \\
Respiratory                    &  -20.68 &    35.98 &  50.12 &   52.36 &  15.30 &  23.19 \\
Oncology                       &  -21.23 &    36.51 &  44.64 &   31.95 &   9.29 &  22.71 \\
Ophthalmology                  &  -12.72 &    16.32 &  33.48 &   37.70 &   4.04 &  14.93 \\
Endocrinology                  &  -23.58 &    42.52 &  41.06 &   51.36 &  18.00 &  41.03 \\
Gastrointestinal               &  -23.73 &    27.60 &  64.51 &   55.60 &  29.60 &   5.21 \\
Hepatology                     &  -20.55 &    30.41 &   7.22 &  124.79 &  18.81 &  22.28 \\
Hematology                     &  -20.08 &    19.99 &  39.54 &   46.01 &  10.31 &  26.03 \\
Rheumatology                   &  -26.09 &    37.06 &  49.10 &   38.39 &   6.39 &  31.13 \\
Immunology                     &  -15.96 &    17.29 &  22.94 &   12.47 &   6.95 &   6.14 \\
Other                          &  -31.31 &    44.95 &  77.03 &   54.13 &  35.01 &  75.59 \\
Orthopedics                    &  -20.50 &    26.17 &   9.87 &   65.54 &  21.97 &  18.64 \\
Psychiatry                     &  -25.04 &    31.42 &  56.57 &   27.68 &  15.27 &  19.59 \\
Women's Health / Sexual Health &  -22.27 &    38.00 &  61.03 &   44.80 &   3.04 &  -0.61 \\
Cardiovascular                 &  -25.62 &    50.83 &  42.83 &   64.75 &  13.58 &  12.04 \\
Infectious Disease             &  -23.66 &    30.13 &  44.35 &   41.25 &  11.93 &  16.26 \\
Dermatology                    &  -20.90 &    22.83 &  66.96 &   43.95 &  13.29 &  32.57 \\
Nephrology                     &  -27.04 &    39.03 &  26.18 &   54.21 &   3.49 &  22.11 \\
\bottomrule
\end{tabular}
\end{center}
\end{table*}

\begin{table*}[h!]
\small
\begin{center}
\caption{Change in the expected patient volume for therapeutic areas with a negative change in representation. Although the relative change for the 'Others' group is negative for one therapeutic area, the change in absolute patient volume is very small. }
\label{tab:neg_TA_perf}
    \begin{tabular}{l|c|c|c}
    \toprule
    Therapeutic area & Others & Pat. volume without div. constraints & Pat. volume with div. constraints\\
    \midrule
Women's health  &-0.61\%& 1222 & 1214\\
\bottomrule
\end{tabular}
\end{center}
\end{table*}

\bibliographystyle{unsrt}
\bibliography{refs}
\end{document}